\documentclass[accepted]{uai2021} 

\usepackage[british]{babel}

\usepackage{natbib} 
\usepackage{mathtools} 
\usepackage{booktabs} 
\usepackage{tikz} 

\usepackage{graphicx}
\usepackage{amsmath}
\usepackage{amsthm}
\usepackage[ruled,vlined]{algorithm2e}
\usepackage{statex2}
\usepackage{dsfont}
\usepackage{mleftright}
\usepackage{comment}
\usepackage{placeins}
\usepackage{pgfplots}
\usepackage{graphicx}
\usepackage[capitalise]{cleveref}

\usepackage[tight-spacing=true]{siunitx}
\newlength\figurewidth

\usetikzlibrary{colorbrewer}

\usepackage{lipsum}

\usepackage{subfiles}
        
\usepackage{rotating}
\usepackage[capitalise]{cleveref}

\usetikzlibrary{positioning}
\usetikzlibrary{shapes.geometric}
\usetikzlibrary{arrows}
\usetikzlibrary{arrows.meta}
\usetikzlibrary{patterns,decorations.pathreplacing}
\usetikzlibrary{fit}
\usetikzlibrary{backgrounds}
\usetikzlibrary{shapes,backgrounds,calc}
\usetikzlibrary{shadows}
\usetikzlibrary{patterns}

\definecolor{color1}{RGB}{141,211,199}
\definecolor{color2}{RGB}{255,255,179}
\definecolor{color3}{RGB}{190,186,218}
\definecolor{color4}{RGB}{251,128,114}
\definecolor{color5}{RGB}{128,177,211}

\newcommand{\cbar}{\,|\,}
\newcommand{\mbf}[1]{\mathbf{#1}}
\newcommand{\indic}[1]{\mathds{1}_{\{#1\}}}
\newcommand{\SumNode}{\mathsf{S}}
\newcommand{\ProductNode}{\mathsf{P}}
\newcommand{\SPN}{\mathfrak{C}}
\newcommand{\Leaf}{\mathsf{L}}

\newcommand{\lse}{\mathrm{L}\overset{K}{\underset{k=1}{\mathrm{\Sigma}}}\mathrm{E}}

\usepackage{xspace}
\newcommand{\eg}{\textit{e.g.}\xspace}
\newcommand{\ie}{\textit{i.e.}\xspace}

\title{Leveraging Probabilistic Circuits for Nonparametric Multi-Output Regression}

\author[1]{\href{mailto:Zhongjie Yu <yu@cs.tu-darmstadt.de>?Subject=Your UAI 2021 paper}{Zhongjie Yu}{}}
\author[2]{Mingye Zhu}
\author[3]{Martin Trapp}
\author[1]{Arseny Skryagin}
\author[1,4]{Kristian Kersting}
\affil[1]{%
    Department of Computer Science\\
    TU Darmstadt\\
    Darmstadt, Germany 
}
\affil[2]{%
    Department of Automation Engineering\\
    Nanjing University of Aeronautics and Astronautics\\
    Nanjing, China
}
\affil[3]{%
    Department of Computer Science\\
    Aalto University\\
    Espoo, Finland
}
\affil[4]{%
    Centre for Cognitive Science\\
    TU Darmstadt\\
    and Hessian Center for AI (hessian.AI)
}
  
\begin{document}
\maketitle

\begin{abstract}
  Inspired by recent advances in the field of expert-based approximations of Gaussian processes (GPs), we present an expert-based approach to large-scale multi-output regression using single-output GP experts. 
  Employing a deeply structured mixture of single-output GPs encoded via a probabilistic circuit allows us to capture correlations between multiple output dimensions accurately. 
  By recursively partitioning the covariate space and the output space, posterior inference in our model reduces to inference on single-output GP experts, which only need to be conditioned on a small subset of the observations. 
  We show that inference can be performed exactly and efficiently in our model, that it can capture correlations between output dimensions and, hence, often outperforms approaches that do not incorporate inter-output correlations, as demonstrated on several data sets in terms of the negative log predictive density. 
\end{abstract}

\section{Introduction}

Gaussian processes (GPs) are a popular class of stochastic processes that can be understood as priors over functions. 
Because of their expressiveness and interpretability---the generalisation properties of a GP are solely determined by choice of the kernel function; they have been heavily used for various machine learning tasks, \eg,~for regression or classification tasks~\citep{RasmussenW06}.
Moreover, GPs have been shown to be closely related to other machine learning models, \eg,~under certain assumptions they correspond to the infinite width limit of Bayesian neural networks~\citep{NealThesis94}.
However, exactly computing the posterior distribution of a GP, \ie,~conditioning a GP prior on $D$-dimensional observations, scales cubic in the number of observations ($N$), \ie,~$\mathcal{O}(N^3)$, and has quadratic memory cost, \ie,~$\mathcal{O}(N^2 + DN)$, thus, limiting their use to moderately sized data sets.

To enable posterior inference in GPs on large-scale problems, recent work (see \eg~\citet{LiuOSC20} for a detailed review) mainly resorts to global approximations to the posterior, \eg,~using inducing points, or local approximations that aim to distribute the computation of the posterior distribution onto local experts.
Unfortunately, most of these approaches only focus on single-output regression, \ie, the dependent variable is univariate, and in the case of local approximations, do not easily extend to multi-output regression tasks, see~\citet{BruinsmaPTHST20} for a detailed discussion on recent techniques on multi-output GPs.

Recently,~\citet{trapp2020deep} proposed a local expert-based approximation to GPs that leverages probabilistic circuits (\eg~\citet{poon2011sum, KisaBCD14, PeharzLVS00BKG20,cho2020pcs}), which are a class of deep tractable probabilistic models, allowing them to perform efficient and exact posterior inference in their model.
In contrast to popular product-of-experts based approaches (\eg~\citet{DeisenrothN15, CohenMMD20}), their method does not approximate the posterior predictive distribution of a GP directly but instead is a model on its own, making it more suitable for further extensions than product-of-experts based approaches.

The contributions of this work are as follows: 
\begin{enumerate}
    \item We propose the multi-output mixture of Gaussian processes (MOMoGP), an extension of~\citet{trapp2020deep} for multi-output regression that scales in $\mathcal{O}(K M^3)$, where $M << N$ is the number of observations per expert, and $K \geq D$ is the number of local experts.
    \item Moreover, we show that posterior inference in our model can be done exactly and reduces to posterior inference at the GP leaves of the networks.
    \item Finally, we present a quantitative evaluation of our approach as well as an application to image upsampling, indicating that MOMoGP is a promising model for multi-output regression.
\end{enumerate}

The rest of the paper is organised as follows. We start by reviewing GP regression and probabilistic circuits in \cref{sec:background}. In \cref{sec:model}, we present MOMoGPs for multi-output regression and discuss posterior inference as well as hyperparameter optimisation. Finally, before concluding, we present a quantitative evaluation of the proposed method in \cref{sec:experiments}.\footnote{Source code is available at: \url{https://github.com/ml-research/MOMoGP}}

\section{Related Work}\label{sec:relatedwork}
{\bf Time Series Regression.} Time series regression is a central task in machine learning. 
We can categorise the existing approach into parametric models, \eg, 
traditional machine learning approaches such as random forest~\citep{breiman2001random}, Adaboost~\citep{solomatine2004adaboost}, XGboost~\citep{chen2016xgboost}, and multi-layer perceptrons~\citep{murtagh1991multilayer}; and nonparametric models, with Gaussian process~\citep{RasmussenW06}, which are distribution over functions, taking the central role in probabilistic machine learning.

{\bf Multi-Output Regression.} 
The methods mentioned above can always be applied to multi-output regression by assuming that all output dimensions are independent. 
However, simply ignoring the correlations among output dimensions will not lead to an accurate representation of the regression task at hand.
One solution is to employ a neural network-based regression model, which can naturally model the output space jointly.
However, accurately quantifying the uncertainties and interpreting the modelled dependencies is often challenging or only possible to a limited extend.

Existing methods for multi-output regression can be categorized into two categories: problem transformation methods and algorithm adaptation methods~\citep{borchani2015survey}.
Problem transformation methods are mainly based on transforming the multi-output regression problem into a single-target problem. 
Consequently, one aggregates the predictions from each single-target regression task to obtain the multi-output predictions. 
Single-Target Method, Multi-Target Regressor Stacking, and Regressor Chains~\citep{spyromitros2012multi} are among problem transformation methods.
Moreover,~\citet{zhang2012multi} presented a multi-output support vector regression approach based on problem transformation, which extends the original feature space and expresses the multi-output problem as an equivalent single-output problem.
On the other hand, algorithm adaptation based methods~\citep{kocev2009using, breiman1997predicting, simila2007input} adapt a specific single-output method to handle multi-output data sets directly. 
These methods generally achieve better results as they consider the underlying relationships between the features and the corresponding targets and the relationships between the targets. 
Existing approaches include reduced-rank regression~\citep{izenman1975reduced,abraham2013position}, multi-output support vector regression~\citep{tuia2011multioutput,xu2013multi}, kernel methods~\citep{baldassarre2012multi,alvarez2011kernels} and multi-target regression trees~\citep{stojanova2012network,appice2014leveraging,levatic2014semi}. 

\citet{williams2007multi} proposed a multi-task Gaussian process, where the model learns a shared covariance function on input-dependent features and a ``free-form'' covariance matrix over tasks.
Further,~\citet{platanios2012nonparametric} presented a nonparametric Bayesian method for multivariate volatility modelling and proposed a mixture of multi-output heteroscedastic GPs to model the covariance matrices of multiple assets. However, this approach is computationally not tractable. 
More recently,~\citet{BruinsmaPTHST20} proposed a linear mixing model, which scales linearly in the number of output dimensions, and showed that their approach could be combined with variational approximations to the GP posterior.

{\bf Probabilistic Circuits for Time Series.} 
Probabilistic circuits (PCs) have previously been used for time series modelling.
\citet{melibari2016dynamic} proposed dynamic sum-product networks for density estimation of time series, which was later extended to so-called recurrent sum-product networks~\citep{kalra2018online, duan2020discriminative} by utilizing discriminative learning.
Recently, \citet{yu2021icml_wspn} proposed to model the distribution of time series in the spectral domain, using Whittle sum-product networks.
While these approaches are able to model the joint distribution of the time series, they often do not allow for straightforward computation of the predictive distribution. 

\cite{trapp2020deep} proposed to define PCs for time series modelling in terms of their induced measure and to equip the PC with Gaussian measures induced by local GPs. 
The resulting model -- called deep structured mixture of Gaussian processes (DSMGP) -- is a deep mixture of na\"ive-local experts.
While DSMGPs aim to approximate GPs by a deep mixture of GP experts, they are limited to single-output GP regression, hence \cite{trapp2020deep} had to resort to a full factorisation for the multi-output regression.

Conversely, MOMoGPs offer a principled way of incorporating multiple output dimensions and model dependencies between outputs through the parameters of the PC. For univariate regression, MOMoGPs reduce to DSMGPs and can therefore be understood as a generalisation of DSMGPs.


\section{Notation and Background}\label{sec:background}
{\bf Notation.} We use the following notations throughout the paper.
$\mathfrak{D}$ is the data set, where $\mathcal X$ is the set of covariates, and $\mathcal Y$ is the set of target values. 
Bold font capitalised $\mbf X$, and $\mbf Y$ are sets of random variables. 
$\mbf X$ is the set of random variables in the covariate space (which is an uncountable set) and $\mbf Y$ the set of output random variables (RVs).
The input space has dimension $D$, the output space has dimension $P$, and the number of observations is $N$.
Furthermore, $\mbf{x}$ denotes the covariates of one observation and $\mbf{y}$ the observed multivariate target/output. 
We use $\mbf{y}_{n}$ to denote the $n^{th}$ observed target/output value from the data set
$\mathfrak{D}$, $y_p$ for the $p^{th}$ dimension in the output space, and $y_{n, p}$ denotes the $p^{th}$ dimension of the $n^{th}$ observed target/output value.

In a GP, as reviewed below, $k(\cdot, \cdot)$ denotes the covariance function and $\mbf{K}$ the covariance (Gram) matrix.
A single-output GP expert is parameterized by hyperparameters $\theta_{\Leaf}$.

In a PC $\mathfrak{C}$, as reviewed in the next section, 
$\SumNode$ represents a sum node, $\ProductNode_x$ represents a product node that partitions the covariate space, 
$\ProductNode_y$ represents a product node that partitions the output space, 
$\Leaf$ is a leaf node, 
and $\mathsf{N}$ denotes a general node.
Furthermore, $K_{\SumNode}$ denotes the number of children of a sum node, and similarly, $K_{\ProductNode_x}$, $K_{\ProductNode_y}$ denote the numbers of children of a product node.
The maximum number of observations per leaf is denoted as $M$.

\subsection{\bf Gaussian Process Regression}
A Gaussian process (GP) is defined as an (uncountable) collection of random variables $\mbf{X}$ indexed by an arbitrary covariate space $\mathbb{R}$, where any finite subset of the RVs is multivariate Gaussian distributed and overlapping finite subsets are marginally consistent~\citep{RasmussenW06}.
Moreover, a GP is fully specified by its mean function $m\colon \mathbb{R}^D \mapsto \mathbb{R}$ and its covariance function $k\colon \mathbb{R}^D \times \mathbb{R}^D \mapsto \mathbb{R}$.
Throughout the paper, we assume a zero-mean function without loss of generality.

Let us assume a data set $\mathfrak{D} = \left \{ (\mathbf{x}_n, y_n) \right \}_{n=1}^N$ consisting of $N$ observations and denote $\mathcal{X} = \left \{ \mathbf{x}_n \right \}_{n=1}^N$ and $\mathcal{Y} = \left \{ y_n \right \}_{n=1}^N$. 
Then the covariance matrix $\mbf{K}_{\mathcal{X},\mathcal{X}}$ is given as: 
\begin{equation}
	\begin{aligned}
	&\mbf{K}_{\mathcal{X},\mathcal{X}} = \\
	&\left[\begin{array}{cccc}
	k\left(\mathbf{x}_{1}, \mathbf{x}_{1}\right) & k\left(\mathbf{x}_{1}, \mathbf{x}_{2}\right) & \cdots & k\left(\mathbf{x}_{1}, \mathbf{x}_{N}\right) \\
	k\left(\mathbf{x}_{2}, \mathbf{x}_{1}\right) & k\left(\mathbf{x}_{2}, \mathbf{x}_{2}\right) & \cdots & k\left(\mathbf{x}_{2}, \mathbf{x}_{N}\right) \\
	\vdots & \vdots & \ddots & \vdots \\
	k\left(\mathbf{x}_{N}, \mathbf{x}_{1}\right) & k\left(\mathbf{x}_{N}, \mathbf{x}_{2}\right) & \cdots & k\left(\mathbf{x}_{N}, \mathbf{x}_{N}\right)
	\end{array}\right] \, .
	\end{aligned}
\end{equation}
In single-output GP regression tasks, and assuming a Gaussian likelihood model, the posterior mean and posterior covariance for a test datum $\mathbf{x}^{*}$ can be obtained in closed-form, \ie, 
\begin{equation}
    m_{\mathfrak{D}}(\mathbf{x}^{*}) = \mbf{K}_{\mathbf{x}^*, \mathcal{X}}\mbf{C}^{-1}\mathcal{Y} \, ,
\end{equation}
and
\begin{equation}
    V_{\mathfrak{D}}(\mathbf{x}^{*}) = \mbf{K}_{\mathbf{x}^*, \mathbf{x}^*} - \mbf{K}_{\mathbf{x}^*, \mathcal{X}}\mbf{C}^{-1}\mbf{K}_{\mathbf{x}^*,\mathcal{X}}^T \, ,
\end{equation}
where $\mbf{C} = [\mbf{K}_{\mathcal{X},\mathcal{X}} + \sigma^2 \mathbf{I}]$, with $\sigma^2$ denoting the noise variance and $\mathbf{I}$ denoting the identity matrix.
However, computing the posterior predictive distribution scales poorly with the number of observations, as the matrix inversion
of $\mbf{C}$ required in the computations has computational cost cubic in $N$, if solved using a Cholesky decomposition.

\subsection{\bf Probabilistic Circuits}
Probabilistic circuits (PCs) are tractable probabilistic models, defined as rooted directed acyclic graphs (DAGs), in which leaf nodes represent univariate probability distributions and non-terminal nodes represent either a mixture (or states of an observed variable in case of a deterministic circuit) or an independence relation of their children. 

More formally, a PC $\mathfrak{C}$ over a set of RVs $\mbf{X}$ is a probabilistic model defined via a DAG, also called the computational graph, containing \emph{input distributions} (\emph{leaves}), \emph{sums} $\mathsf{S}$ and \emph{products} $\mathsf{P}$; and a scope function $\textbf{sc}(\cdot)$.
We refer to~\citet{TrappPGPG19} for further details.

For a given scope function, all leaves of the PC are density functions over some subset of RVs $\mbf{U} \subset \mbf{X}$. 
This subset is called the scope of the node, and for a node $\mathsf{N}$ is denoted as $\textbf{sc}(\mathsf{N})$.
The scope of inner nodes is defined as the union of the scope of its children.
Inner nodes compute either a weighted sum of their children or a product of their children, \ie,~$\mathsf{S} = \sum_{\mathsf{N}\in \textbf{ch}(\mathsf{S})} w_{\mathsf{S},\mathsf{N}}\mathsf{N}$ 
and $\mathsf{P} = \prod_{\mathsf{N}\in \textbf{ch}(\mathsf{P})} \mathsf{N}$, where $\textbf{ch}(\cdot)$ denotes the children of a node.
The sum weights $w_{\mathsf{S},\mathsf{N}}$ are assumed to be non-negative and normalized, \ie,~$w_{\mathsf{S},\mathsf{N}} \geq 0, \sum_{\mathsf{N}}w_{\mathsf{S},\mathsf{N}}=1$, without loss of generality~\citep{PeharzTPD15}.
Further, we assume the PC to be smooth (complete) and decomposable~\citep{darwiche2003differential}.
Specifically, a PC is \emph{smooth} if for each sum $\mathsf{S}$ it holds that $\textbf{sc}(\mathsf{N'})=\textbf{sc}(\mathsf{N''})$, for all $\mathsf{N'}, \mathsf{N''} \in \textbf{ch}(\mathsf{S})$.
And the PC is called \emph{decomposable} if it holds for each product $\mathsf{P}$ that $\textbf{sc}(\mathsf{N'})\cap \textbf{sc}(\mathsf{N''})= \emptyset$, for all $\mathsf{N'} \neq \mathsf{N''} \in \textbf{ch}(\mathsf{P})$.

Note that PCs are typically defined only for a finite set of RVs, while~\citet{trapp2020deep} showed that it is possible to extend PCs to the stochastic process case by defining them based on their induced measure. We will, therefore, follow the approach in~\citep{trapp2020deep} and recursively define our model as such.

\section{Multi-Output Mixture of Gaussian Processes}\label{sec:model}
Now we have everything at hand to introduce our MOMoGP model formally. 
We start off with the problem formulation, introduce our MOMoGP model, and subsequently show how to perform exact posterior inference as well as how to perform predictions using MOMoGPs. 
Finally, we discuss hyperparameter optimisation in MOMoGPs by maximising the marginal likelihood.

\subsection{Problem Formulation}\label{sec:problemformulation}
Given a set of observations $\mathfrak{D} = \{ ({\bm x_n},{\bm y_n}) \}^N_{n=1}$ with covariates ${\bm x_n} \in \mathbb{R}^D$ and noisy target values ${\bm y_n} \in \mathbb{R}^P$, we aim to infer the latent functions:
\begin{align}
    f_{p} &\sim \text{GP}(\bm 0, \mbf{K}) \, , \\
    y_{p} \cbar f_{p} &\sim N(f_{p}(\bm x), \sigma^2\mbf{I}) \, ,
\end{align}
while aiming to account for the correlations between the target values.
One approach is to model the multi-output targets by adopting a multi-valued latent function. However, such an approach scales cubic in the number of output dimensions $P$, \ie,~$\mathcal{O}(N^3P^3)$~\citep{BruinsmaPTHST20}.
Alternatively, we exploit a simple observation, that is, we can leverage a mixture of independent GP estimators to capture correlations between the output dimensions. 
This is akin to the Instantaneous Linear Mixing Model by~\citet{BruinsmaPTHST20} but explores recent work by~\citet{trapp2020deep} to perform efficient and exact posterior inference in a deep mixture over single-output GP experts to obtain correlated multi-output predictions.
Note that the approach by \citet{BruinsmaPTHST20} assumes that the outputs live on a low-dimensional linear subspace and exploits an orthogonal basis for efficient inference, while our approach assumes weak independence between output dimensions and exploits the multi-modality of the posterior to capture dependencies between output dimensions.

\subsection{Multi-Output Mixture of GPs}

The multi-output mixture of Gaussian processes \mbox{(MOMoGP)} can be recursively defined as follows:
\begin{enumerate}
  \item A Gaussian measure induced by a GP \citep{Rajput1972} is a MOMoGP,
  \item a product of MOMoGPs with disjoint covariate space or disjoint output space is a MOMoGP, and
  \item a convex combination of MOMoGPs over the same covariate and output space is a MOMoGP.
\end{enumerate}

\pgfdeclarelayer{background}
\pgfdeclarelayer{foreground}
\pgfsetlayers{background,main,foreground}

\begin{figure}[thpb]
\centering
\begin{tikzpicture}[>=latex, minimum size=8mm, inner sep=0pt, align=center]
\node[draw=black!75, fill=gray!20, thick, circle, label={[align=center]right: $\SumNode$}] (s) at (0,-1.5) {\large$+$};
\node[] (pdots) at (-2.5,-4) {\large$\dots$};
\node[] (p2) at (2.5,-4) {\large$\dots$};
\node[draw=black!75, fill=gray!20, thick, circle, label={[align=center]right: $\ProductNode_{x}$}] (p1) at (0,-4) {\large$\times$};
\node[draw=black!75, fill=gray!20, thick, circle, label={[align=center]right: $\ProductNode_{y}$}] (pro1) at (0,-6.5){\large$\times$};
\node[] (pro0) at (-2.5,-6.5) {\large$\dots$};
\node[] (pro2) at (2.5,-6.5) {\large$\dots$};
\node[draw=black!75, fill=gray!20, thick, circle] (s21) at (0,-9) {GP};
\node[] (s20) at (-2.5,-9) {\large$\dots$};
\node[] (s22) at (2.5,-9) {\large$\dots$};
\draw[->,thick] (s) -- (pdots)
    node[midway,fill=white] {$w_{1,1}$};
\draw[->,thick] (s) -- (p1)
    node[midway,fill=white] {$w_{1,k}$};
\draw[->,thick] (s) -- (p2)
    node[midway,fill=white] {$w_{1,K_{\SumNode}}$};
\draw[->,thick] (p1) -- (pro0)
    node[midway,fill=white] {$\mathbf{X}_{t < i}$};
\draw[->,thick] (p1) -- (pro1) 
    node[midway,fill=white] {$\mbf{X}_{i \leq t < j}$};
\draw[->,thick] (p1) -- (pro2) 
    node[midway,fill=white] {$\mbf{X}_{t \geq j}$};
\draw[->,thick] (pro1) -- (s20)
    node[midway,fill=white] {$\mbf{Y}^{l}_1$};
\draw[->,thick] (pro1) -- (s21) 
    node[midway,fill=white] {$Y_{l+1}$};
\draw[->,thick] (pro1) -- (s22) 
    node[midway,fill=white] {$\mbf{Y}^P_{l+2}$};
\end{tikzpicture}
\caption{Illustration of the MOMoGP structure. $w_{1,k}$ represents for the normalized weight, $\mbf{X}_{i \leq t < j} \subset \mbf{X}$ represents the subset of RVs $X_t$ with index $i \leq t < j$ of the covariate space, and $\mbf{Y}^{l}_1$ denotes a subset of RVs of the output space, respectively. Note that we randomly permute the index set of the output space at each product node $\ProductNode_{y}$.}
\label{fig:MOMoGP}
\end{figure}
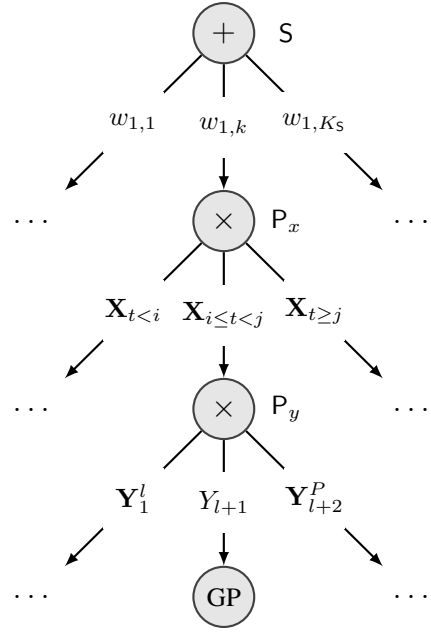

The recursive definition of a MOMoGP is illustrated in \cref{fig:MOMoGP}. Here, $w_{1,1}, \cdots , w_{1,k}, \cdots, w_{1,K_{\SumNode}}$ are $K_{\SumNode}$ many normalized weights of the sum node $\SumNode$.
The product node $\ProductNode_x$, which is a child of the root, splits the (input) covariate space $\mbf{X}$, by assuming certain regions of the input space to be approximately independent.
The second product node, $\ProductNode_y$, partitions the output space $\mbf{Y}$ into disjoint subsets.
More specifically, its children are either MOMoGPs over a set of output variables, \eg,~$\mbf{Y}_1^l$, or only a single variable, \eg,~$Y_{l+1}$.
For a univariate output, \eg,~$Y_{l+1}$ we construct a GP leaf $\Leaf$ on the respective covariate subspace and output subspace.
Otherwise, the process recurses by appending a new sum node, resulting in a deep hierarchical structure.

\subsection{MOMoGP Construction}\label{sec:structure}
The structure of a MOMoGP can either be manually defined or learned from data.
To learn it from data, one can leverage \cref{alg:structure}. In short, we alternate between introducing sum and product nodes and, finally, append GP experts as leaves once any of the termination criteria is fulfilled. 

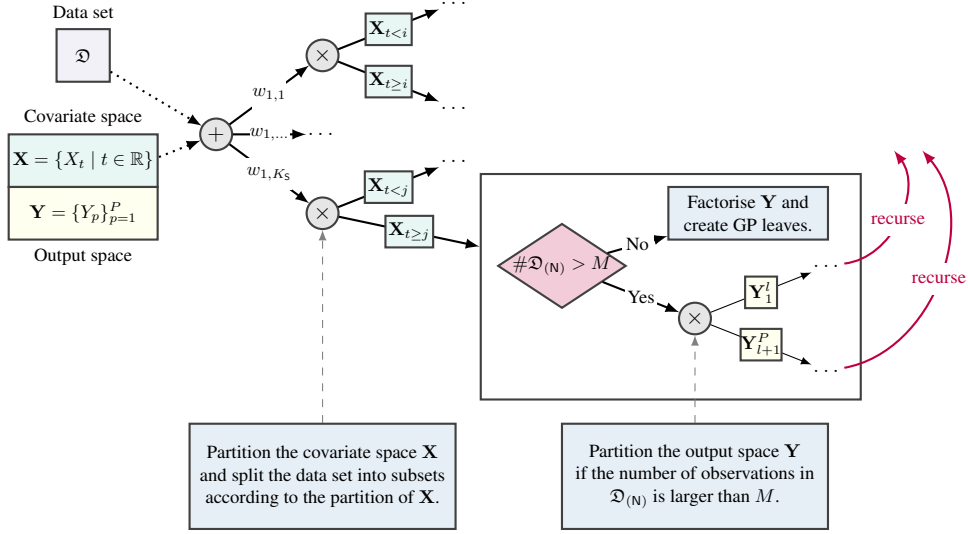
\begin{figure*}[thpb]
      \centering
      \begin{tikzpicture}[>=latex, minimum size=6mm, inner sep=0pt, align=center]
      \begin{scope}[scale=0.7, transform shape]
        \node[draw=black!75, fill=color1!20, thick, minimum height=1cm, minimum width=2.8cm, label={[align=center]above:Covariate space}] (xin) at (0.5,-2) {$\mathbf{X} = \{X_t \mid t \in \mathbb{R}\}$};
        \node[draw=black!75, fill=color2!20, thick, minimum height=1cm, minimum width=2.8cm, label={[align=center]below:Output space}] (yin) at (0.5,-3) {$\mathbf{Y} = \{Y_p\}^P_{p=1}$};
        \node[fit=(yin)(xin)](group){};
        \node[draw=black!75, fill=color3!20, thick, minimum height=1cm, minimum width=1cm, label={[align=center]above:Data set}, anchor=center](D) at (0.5, 0) {$\mathfrak{D}$};

        \node[draw=black!75, fill=gray!20, thick, circle] (RootN) at (3,-1.5) {\large$+$};
        \node[draw=black!75, fill=gray!20, thick, circle] (SplitN1) at (5, 0) {\large$\times$};
        \node[] (SplitN2) at (5,-1.5) {\large$\dots$};
        \node[draw=black!75, fill=gray!20, thick, circle] (SplitN3) at (5, -3) {\large$\times$};
        
        \node[](pro11) at (7.5,1.0) {\large$\dots$};
        \node[] (pro12) at (7.5,-1){\large$\dots$};

        \node[draw=black!75, fill=gray!20, thick, circle] (pro21y) at (12,-5) {\large$\times$};
        
        \node[] (hy1) at (14.5,-4) {\large$\dots$};
        \node[] (hrecurse1) at (15.5,-1.5) {};
        \node[] (hy2) at (14.5,-6) {\large$\dots$};
		\node[] (hrecurse2) at (16,-1.5) {};
        
        \node[draw=black!75, fill=purple!20, thick, minimum width=2cm, diamond, aspect=1.5] (WL) at (9.5,-4){$\#\mathfrak{D}_{(\mathsf{N})}>M$};

        
        \node[draw=black!75, fill=color5!20, thick, minimum height=1cm, minimum width=3cm, rectangle] (WL3) at (13,-3){Factorise $\mathbf{Y}$ and\\ create GP leaves.};

		\node[fit=(WL)(pro21y)(WL3)(hy1)(hy2), draw=black!75, thick, rectangle, minimum height=3cm, minimum width=5cm](group2){};
		
        \node[] (pro22) at (7.5,-2) {\large$\dots$};



        \node[draw=black!75, fill=color5!20, thick, minimum width=5cm, minimum height=2cm, rectangle] (WL1) at (5,-8){Partition the covariate space $\mathbf{X}$\\ and split the data set into subsets\\ according to the partition of $\mathbf{X}$.};
        \node[draw=black!75, fill=color5!20, thick, minimum height=2cm, minimum width=5cm, rectangle] (WL2) at (12,-8){Partition the output space $\mathbf{Y}$\\ if the number of observations in\\ $\mathfrak{D}_{(\mathsf{N})}$ is larger than $M$.};

        \draw[->,thick, bend right, anchor=north east, dotted] (group) -- (RootN);
        \draw[->,thick, bend right, anchor=south east, dotted] (D) -- (RootN);
        \draw[->,thick] (RootN) -- (SplitN1) node[midway,fill=white] {$w_{1,1}$};
        \draw[->,thick] (RootN) -- (SplitN2) node[midway,fill=white] {$w_{1,\dots}$};
        \draw[->,thick] (RootN) -- (SplitN3) node[midway,fill=white] {$w_{1,K_{\SumNode}}$};
        \draw[->,thick] (SplitN1) -- (pro11) node[midway, fill=color1!20, draw=black!75, thick] {$\,\mbf{X}_{t < i}\,$};
        \draw[->,thick] (SplitN1) -- (pro12) node[midway, fill=color1!20, draw=black!75, thick] {$\,\mbf{X}_{t \geq i}\,$};
        
        
        \draw[->] (pro21y) -- (hy1) node[midway, fill=color2!20, draw=black!75, thick] {$\mbf{Y}_1^l$};
        \draw[->] (pro21y) -- (hy2) node[midway, fill=color2!20, draw=black!75, thick] {$\mbf{Y}_{l+1}^P$};
        
        \draw[->,thick] (SplitN3) -- (pro22) node[midway, fill=color1!20, draw=black!75, thick] {$\,\mbf{X}_{t<j}\,$};
        \draw[->,thick] (SplitN3) -- (group2) node[midway, fill=color1!20, draw=black!75, thick] {$\,\mbf{X}_{t \geq j}\,$};
        \draw[->,thick] (WL) -- (WL3) node[midway,fill=white] {No};
        \draw[->,thick] (WL) -- (pro21y) node[midway,fill=white] {Yes};
        \draw[->,dashed,gray] (WL2) -- (pro21y);
        \draw[->,dashed,gray] (WL1) -- (SplitN3);
        
        
        \draw[->,purple, thick] (hy1) edge[bend right=60] node[midway, fill=white]{recurse} (hrecurse1);
        \draw[->,purple, thick] (hy2) edge[bend right=60] node[midway, fill=white]{recurse} (hrecurse2);

    \end{scope}
    \end{tikzpicture}
    \caption{Illustration of \cref{alg:structure} for learning MOMoGPs in a recursive fashion. Sum nodes have weighted children that are product nodes. Product nodes either enforce independence assumptions in the covariate space (resulting in discontinuities) or independence assumptions in the output space (assuming sets of the dependent variables are independent). Sum nodes replace the independence assumptions made by product nodes through conditional independence. Finally, leaf nodes are single-output GP experts at the respective covariate subspace. (Best viewed in color) \label{learnMOMoGP}}
\end{figure*}

As illustrated in \cref{learnMOMoGP}, to create a sum node $\mathsf{S}$, we first construct $K_{\SumNode}$ many children under the sum and then attach those children with uniform weights.
In the next step, a product node $\ProductNode_x$ is constructed by partitioning the covariate space. 
For the $k^{th}$ product node, the covariate space is partitioned by the $K_{\ProductNode_x}$-quantiles of the dimension with the $k^{th}$ largest sample variance.
Afterwards, a product node $\ProductNode_y$ is created by randomly partitioning the output space. 
To split the output space one can also apply a conditional independence test on $\mathbf{y}$ instead of random splitting. 
Product nodes either enforce independence assumptions in the covariate space (resulting in weak discontinuities) or independence assumptions in the output space (assuming sets of the dependent variables are conditionally independent).
The above sum and product nodes are constructed recursively until the number of observations in the subspace is smaller than a predefined threshold $M$.
If the output space is still multidimensional, it will be directly factorized with a product node $\ProductNode_y$.
Finally, we construct leaf nodes by placing single-output GP experts at the respective covariate subspace parameterized by hyperparameters $\mathsf{\theta _L}$.


\begin{algorithm}[t!]
    \SetKwFunction{buildGP}{buildGP}
    \SetKwFunction{buildSumNode}{buildSumNode}
    \SetKwFunction{buildProductNodeX}{buildProdNodeX}
    \SetKwFunction{buildProductNodeY}{buildProdNodeY}
    \KwIn{$\mbf{X}, \mbf{Y}, \mathfrak{D}$, $K_{\SumNode}$, $K_{\ProductNode x}$, $K_{\ProductNode y}$, $M$ \quad \textbf{Output:} \ $\SPN$ }

    \SetKwProg{Fn}{Function}{}{end}
    \Fn{\buildGP{$\mbf{X}, Y, \mathfrak{D}$}}{
        Equip $\Leaf$ with a single output GP expert on the domain $\mbf{X}$ and output space $Y$ \; 
        Condition $\Leaf$ on $\mathfrak{D}$ \;
        \Return{$\Leaf$}
    }

    \Fn{\buildSumNode{$\mbf{X}, \mbf{Y}, \mathfrak{D}$}}{
        $w \gets \{\frac{1}{K_\SumNode}\}^{K_\SumNode}_{k=1}$\;
        Let $d_1, \dots, d_D$ represent the dimensions of the covariate space in increasing order of their sample variance in $\mathfrak{D}$\;
        \For{$k=1, \dots, K_\SumNode$}{
            $\ProductNode_x \gets$ \buildProductNodeX{$\mbf{X}, \mbf{Y}, \mathfrak{D}, d_k$}\;
            $\SumNode \gets \SumNode + w_k\, \ProductNode_x$\;
        }
        \Return{$\SumNode$}
    }

    \Fn{\buildProductNodeX{$\mbf{X}, \mbf{Y}, \mathfrak{D}, d$}}{
        $l \gets $ lower bound of $\mbf{X}$ for dimension $d$\;
        $u \gets $ upper bound of $\mbf{X}$ for dimension $d$\;
        $v \gets u-l$ \;
        $s_1, \dots, s_{K_{\ProductNode x}-1} \gets $ $K_{\ProductNode x}$-quantiles of the interval [$l, u$]\;
        $\tilde{l} \gets l$ \;
        \For{$k=1, \dots, K_{\ProductNode x}-1$}{
            $\tilde{u} \gets s_k$ \;
            $\mbf{\tilde{X}} \gets $ sub-domain of $\mbf{X}$ such that upper and lower bounds for $d$ are equal to $\tilde{u}, \tilde{l}$, respectively. \;
            $\mathcal{\tilde{\mathfrak{D}}} \gets $ subset of $\mathfrak{D}$ such that the covariate of every $(\bm x_n, \bm y_n) \in \mathcal{\tilde{\mathfrak{D}}}$ is defined on $\mbf{\tilde{X}}$. \;
            $\ProductNode_y \gets$ \buildProductNodeY{$\mbf{\tilde{X}}, \mbf{Y}, \mathcal{\tilde{\mathfrak{D}}}$}\;
            $\ProductNode \gets \ProductNode \times \ProductNode_y$ \; 
            $\tilde{l} \gets s_k$ \;
        }
        \Return{$\ProductNode$}
    }
    
    \Fn{\buildProductNodeY{$\mbf{X}, \mbf{Y}, \mathfrak{D}$}}{
        \eIf{Number of observations in $\mathfrak{D} > M$}{
            $\mbf{Y}_1, \cdots , \mbf{Y}_{K_{\ProductNode y}-1} \gets$ random partitions of output space $\mbf{Y}$\;
            \For{$k=1, \dots, K_{\ProductNode y}-1$}{
                $\SumNode \gets $ \buildSumNode{$\mbf{X}, \mbf{Y}_k, \mathfrak{D}$} \;
                $\ProductNode \gets \ProductNode \times \SumNode$ \;
            }
        }{
            $Y_1, \cdots , Y_{K_{\ProductNode y}-1} \gets$ each dimension of output space $\mbf{Y}$\;
            \For{$k=1, \dots, K_{\ProductNode y}-1$}{
                $\Leaf \gets $ \buildGP{$\mbf{X}, Y_k, \mathfrak{D}$} \;
                $\ProductNode \gets \ProductNode \times \Leaf$  \;
            }
        }
                
        \Return{$\ProductNode$}
    }
    $\SPN \leftarrow$ \buildSumNode{$\mbf{X}, \mbf{Y}, \mathfrak{D}$} \;
\caption{Construction of a MOMoGP \label{alg:structure}} 
\label{alg:MOMoGP}
\end{algorithm}

\subsection{Exact Posterior Inference}\label{sec:posterior}
Both PCs and GPs allow for exact posterior inference, likewise, we can formalize the exact posterior inference for MOMoGP as follows:

\paragraph{(Leaf node)} If the MOMoGP is a leaf node $\mathsf{L}$ we can obtain the posterior distribution analytically, assuming a Gaussian likelihood.

\paragraph{(Sum node)} If the MOMoGP is a sum node $\mathsf{S}$, the likelihood terms distribute to the children, \ie, 
\begin{equation}
	\begin{aligned}
   p_{\mathsf{S}}(f \mid \mathfrak{D}) & \propto \prod_{\mathclap{\left(\mathbf{x}_{n}, \mathbf{y}_{n}\right) \in \mathfrak{D}}} \, p\left(\mathbf{y}_{n} \mid f_{n}\right) \sum_{\mathclap{\mathsf{N} \in \operatorname{ch}(\mathsf{S})}} \, w_{\mathsf{S}, \mathsf{N}} \, p_{\mathsf{N}}\left(f_{n} \mid \mathbf{x}_{n}\right)\\
   & =\sum_{\mathclap{\mathsf{N} \in \operatorname{ch}(\mathsf{S})}} w_{\mathsf{S}, \mathsf{N}} \, \underbrace{\prod_{\mathclap{\left(\mathbf{x}_{n}, \mathbf{y}_{n}\right) \in \mathfrak{D}}} \, p\left(\mathbf{y}_{n} \mid f_{n}\right) \, p_{\mathsf{N}}\left(f_{n} \mid \mathbf{x}_{n}\right)}_{= p_{\mathsf{N}}(f \mid \mathfrak{D})} \, ,
   \end{aligned}
\end{equation}
simplifying the computation of the unnormalised posterior distribution. Note that instead of being univariate as in~\citet{trapp2020deep}, $\mathbf{y}_{n}$ is multidimensional in MOMoGP.

\paragraph{(Product node)} If the MOMoGP is a product node $\mathsf{P}$ decomposing either the covariate space or the output space, we obtain:
\begin{equation}
	\begin{aligned}
p_{\mathsf{P}}(f \mid \mathfrak{D}) &\propto \prod_{\mathclap{\left(\mathbf{x}_{n}, \mathbf{y}_{n}\right) \in \mathfrak{D}}} \, p\left(\mathbf{y}_{n} \mid f_{n}\right) \prod_{\mathclap{\mathsf{N} \in \operatorname{ch}(\mathsf{P})}} \, p_{\mathsf{N}}\left(f_{n} \mid \mathbf{x}_{n}\right) \\
&=\prod_{\mathclap{\mathsf{N} \in \operatorname{ch}(\mathsf{P})}} \,\,\, \Bigg( \;\;\;\;\;\; \underbrace{\prod_{\mathclap{\left(\mathbf{x}_{n}, \mathbf{y}_{n}\right) \in \mathfrak{D}_{(\mathsf{N})}}} \, p\left(\mathbf{y}_{n} \mid f_{n}\right) \, p_{\mathsf{N}}\left(f_{n} \mid \mathbf{x}_{n}\right)}_{= p_{\mathsf{N}}(f \mid \mathfrak{D}_{(\mathsf{N})})} \Bigg)\, ,
   \end{aligned}
\end{equation}
for the computation of the unnormalised posterior, where $\mathfrak{D}_{(\mathsf{N})}$ denotes the set of observations in the subspace for which $\mathsf{N}$ is an expert of. 
In the case of the covariate space decomposition ($\ProductNode_x$), $\mathfrak{D}_{(\mathsf{N})} = \left \{ \left ( \mathbf{x}_n, \mathbf{y}_n \right ) \in \mathfrak{D} \cbar \mathbf{x}_n \in \mathcal{X}_\mathsf{N} \right \}$, where $\mathcal X_{\mathsf{N}}$ is the subset of covariates falling into the subspace at node $\mathsf{N}$.
While for the output space decomposition ($\ProductNode_y$), $\mathfrak{D}_{(\mathsf{N})}$ contains a subset of outputs for each observation, \ie, $\mathfrak{D}_{(\mathsf{N})} = \left \{ \left ( \mathbf{x}_n, \mathbf{y}_n \right ) \in \mathfrak{D} \cbar \mathbf{y}_n \in \mathcal{Y}_\mathsf{N}^{P_{\mathsf{N}}}, \mathcal{Y}_\mathsf{N}^{P_{\mathsf{N}}} \subseteq \mathcal{Y}_\mathsf{N}, P_\mathsf{N} < P \right \}$. 
Therefore, $\mathfrak{D}_{(\mathsf{N})}$ will either contain fewer observations, \ie,~$\#\mathfrak{D}_{(\mathsf{N})} < \#\mathfrak{D}$, or fewer output dimensions, \ie,~for each $n$ we have $\mathbf{y}_{n} \in \mathfrak{D}_{(\mathsf{N})}$ with $\mathbf{y}_{n} \in \mathbb{R}^{P_{\mathsf{N}}}$ and ${P_{\mathsf{N}}} < P$. 
In contrast to \citet{trapp2020deep}, product nodes in MOMoGPs partition either the covariate space or the output space, while DSMGPs \citep{trapp2020deep} only partition the covariate space and assume the output space to be univariate.

To obtain the normalised posterior distribution of a MOMoGPs we employ the re-normalisation algorithm by \cite{PeharzTPD15}.

\subsection{Predictions}
The posterior predictive distribution of a MOMoGP for an unseen datum $\mathbf{x}^*$ is a mixture of multivariate Gaussian distributions.
To obtain a single prediction for $\mathbf{x}^*$, we employ an approximation to the posterior predictive distribution of the MOMoGP using the first and the second moment.
By this, we approximate the posterior predictive distribution with its closest multivariate Gaussian measured in Kullback–Leibler divergence~\citep{RasmussenW06}. 

Let $\mathfrak{L}$ be the set of all leaves in a MOMoGP, and $\tau_i \colon \mathbb{R}^D \mapsto \mathfrak{L}$ a function which maps an unseen datum $\mathbf{x}^*$ to a leaf $\mathsf{L}$ for each induced tree. 
Then the posterior mean and variance given $\mathbf{x}^*$, are propagated bottom-up as follows. 
Let $m_{\tau_i (\mathbf{x}^*)}(\mathbf{x}^*)$ and $V_{\tau_i (\mathbf{x}^*)}(\mathbf{x}^*)$ denote the mean and variance of the posterior distribution from the GP at leaf $\tau_i (\mathbf{x}^*)$, respectively. 
Now, a product node $\ProductNode_y$ that partitions the output space performs concatenation of the mean and variance from its children:
\begin{align}
    m_{\mathsf{P}}(\mathbf{x}^*) &= [m_{\mathsf{N}_1}(\mathbf{x}^*), \cdots , m_{\mathsf{N}_k}(\mathbf{x}^*)] \, , \\
    \text{and} \ V_{\mathsf{P}}(\mathbf{x}^*) &= \text{diag}(V_{\mathsf{N}_1}(\mathbf{x}^*) \, , \cdots , V_{\mathsf{N}_k}(\mathbf{x}^*)) \; .
\end{align}
In contrast, a product node $\ProductNode_x$ that partitions the covariate space acts as a gate, \ie, we select exactly one child of the product node. The selection of the child is dictated by the partition of covariate space employed by the product node and the location of $\mathbf{x}^*$.
For a sum node $\mathsf{S}$, it computes the first moment, \ie,
\begin{equation}
    m_{\mathsf{S}}(\mathbf{x}^*) = \sum_{\mathclap{\mathsf{N} \in \textbf{ch}(\mathsf{S})}} \, w_{\mathsf{S}, \mathsf{N}} \, m_{\mathsf{N}}(\mathbf{x}^*) \, ,
\end{equation}
and the second moment ,\ie, 
\begin{equation}
	\begin{aligned}
    V_{\mathsf{S}}(\mathbf{x}^*) &= \sum_{\mathclap{\mathsf{N} \in \textbf{ch}(\mathsf{S})}} \, w_{\mathsf{S}, \, \mathsf{N}} V_{\mathsf{N}}(\mathbf{x}^*) 
    + \sum_{\mathclap{\mathsf{N} \in \textbf{ch}(\mathsf{S})}} \, w_{\mathsf{S}, \mathsf{N}} \,  m_{\mathsf{N}}(\mathbf{x}^*)^{T} \, m_{\mathsf{N}}(\mathbf{x}^*) \\
    &- m_{\mathsf{S}}(\mathbf{x}^*)^{T}\, m_{\mathsf{S}}(\mathbf{x}^*) \, ,
    \end{aligned}
\end{equation}
of the associated mixture distribution.

\subsection{Hyperparameter Optimisation}
To optimise the hyperparameters of a MOMoGP model, we can maximize its log marginal likelihood.
When using a formulation in terms of a mixture over induced trees $\mathcal{T}$ and following the argument from \cref{sec:posterior}, we can see that the marginal likelihood of a MOMoGP is obtained as follows:
\begin{equation}
	\begin{aligned}
&p(\mathcal{Y} \cbar \mathcal{X}, \theta) \\
&= \int \prod_{(\mathbf{x}, \mathbf{y}) \in \mathfrak{D}} \sum^K_{k=1} p(\mathcal{T}_k) \prod^P_{p=1} \prod_{\Leaf \in \mathcal{T}_{k,V}} p(y_p \cbar \mathbf{x}, \theta_\Leaf)\indic{Y_p,\Leaf} \d{\mathbf{f}} \\
&= \int  \sum^K_{k=1} p(\mathcal{T}_k) \prod^P_{p=1} \prod_{\Leaf \in \mathcal{T}_{k,V}} \prod_{(\mathbf{x}, y_p) \in \mathfrak{D}_{(\Leaf)}} p(y_p \cbar \mathbf{x}, \theta_\Leaf)\indic{Y_p,\Leaf} \d{\mathbf{f}} \\
&= \sum^K_{k=1} p(\mathcal{T}_k) \prod^P_{p=1} \prod_{\Leaf \in \mathcal{T}_{k,V}} \underbrace{\int \prod_{(\mathbf{x}, y_p) \in \mathfrak{D}_{(\Leaf)}} p(y_p \cbar \mathbf{x}, \theta_\Leaf)\indic{Y_p,\Leaf} \d{\mathbf{f}}}_{=p_{Y_p}(y_p \cbar \mathcal{X}_\Leaf, \theta_\Leaf)}\, ,
	\end{aligned}
\end{equation}
where $\mathfrak{D}_{(\Leaf)} = \left \{ \left ( \mathbf{x}, y_p \right ) \in \mathfrak{D} \cbar \mathbf{x} \in \mathcal{X}_\Leaf \right \}$, 
with $\mathcal X_\Leaf$ being the subset of covariates falling into the subspace at leaf $\Leaf$, 
and $\indic{Y_p,\Leaf} = \indic{Y_p \in \textbf{sc}(\Leaf)}$. 

We can now readily obtain the log marginal likelihood, which is given as: $\log p(\mathcal{Y} \cbar \mathcal{X}, \theta) =$
\begin{equation}
\lse \mleft( \log p(\mathcal{T}_k) +\sum^P_{p=1} \sum_{\Leaf \in \mathcal{T}_{k,V}}\!\! \log p_{Y_p}(y_p \cbar \mathcal{X}_\Leaf, \theta_\Leaf)\mright) \, ,
\end{equation}
where $\lse$ denotes the log-sum-exp operation and the marginal log likelihood of a GP expert is given as:
\begin{equation}
    \log p_{Y_p}(y_p \mid \mathcal{X}, \theta)=-\frac{1}{2} (y_p^{\mathrm{T}} \mbf{C}^{-1} y_p + \log |\mbf{C}| +N \log 2 \pi) \, ,
\end{equation}
where $\mbf{C} = [\mbf{K}_{\mathcal{X},\mathcal{X}} + \sigma^2 \mathbf{I}]$ and $N$ is the number of observations the GP expert is conditioned on.

Note that we have assumed that each GP expert has its own hyperparameters $\theta_\Leaf$. 
Doing so allows us to capture non-stationarities and heteroscedasticity, while potentially increasing the risk of overfitting~\citep{hangW19}.

\begin{table}[t!]
\centering
\begin{tabular}{lrrrr}\toprule
Data set & $N$(train) & $N$(test) & $D$ & $P$ \\ \midrule
Parkinsons & 4,112 &1,763 & 16 & 2 \\
scm20d & 7,173 & 1,793 & 61 & 16 \\
WindTurbine & 4,000 & 1,000 & 8 & 6 \\
Energy & 57,598 & 14,400 & 32 & 17 \\
usFlight & 500,000 & 200,000 & 8 & 2 \\
\bottomrule
\end{tabular}
\caption{Statistics of the multi-output benchmark data sets used in our evaluation. A large variety of output dimensions $P$ were chosen, ranging from $2$ to $17$. \label{tab:datasets}}
\end{table}

If the underlying process is believed to be stationary, it is possible to tie the hyperparameters either by using one set of global hyperparameters for each output dimension or by adopting the approach described in~\citet{trapp2020deep} and use a similarity matrix to incorporate dependence between the otherwise independent local experts.
Note that we consistently used independent hyperparameters for each GP expert in our experiments.

\section{Experimental Evaluation}\label{sec:experiments}
In this section, we will examine the performance of MOMoGP on several benchmark data sets and compare it with other state-of-the-art approaches.
First, we describe the data sets and then provide details about the experimental setup and evaluation measures we used. 
Finally, we discuss the experimental results obtained.

\subsection{Data Sets}
We validate our model on several benchmark data sets for multi-output regression.
The number of observations in the data sets varies from 4k to 500k, with output space dimensions from $2$ to $17$.
The statistics of the data sets used in the evaluation are given in \cref{tab:datasets}.
The Parkinsons and usFlight data sets, as well as their training/test splits, are from~\citet{trapp2020deep}. 
Note that we applied Principal Component Analysis (PCA)~\citep{wold1987principal} on the scm20d data set to reduce its input dimension from $61$ to $30$.
Both MOMoGPs and DSMGPs recursively partition the covariate space using axis-aligned splits. Therefore, applying PCA to high-dimensional covariate spaces is essential for computational reasons for both approaches.
For the Energy data set, we select its subset ``Adelaide'' for our experiments.
The WindTurbine data set is simulated with the FAST simulator.\footnote{\url{https://www.nrel.gov/wind/nwtc/fast.html}}

\subsection{Experimental Protocol}
To construct a MOMoGP for each experiment, we implemented \cref{alg:MOMoGP}. 
More specifically, the root node of our hierarchy structure was a sum node, with $K_{\SumNode}$ product nodes as children, initialized with uniform weights. 
Product nodes that split the input space had $K_{\ProductNode_x}$ children, and those that decompose the output space used a random split strategy to obtain $K_{\ProductNode_y}$ children. 
The structure construction terminated with GP leaves when the output space was completely decomposed and the number of observations in the subspace is smaller than a predefined threshold of $M$. 

\begin{table*}
\centering
\begin{tabular}{ll|ccccccc} \toprule
\multicolumn{2}{c}{Data Set} & LR     & GP     & MOGP   & MOSVGP   & DSMGP & sumGP  & \textbf{MOMoGP}  \\ \midrule
                 & RMSE     & 0.974  & 0.783  & 0.793  & 0.864  & \textbf{0.774}      & 0.784  & 0.775$\downarrow$  \\
Parkinsons       & MAE      & 0.816  & 0.610  & \textbf{0.603}  & 0.708  & 0.604      & 0.610  & 0.605$\downarrow$  \\
                 & NLPD     & 2.787  & 2.389  & \textbf{1.766}  & 2.515  & 2.319      & 2.388  & 2.208$\uparrow$  \\[0.5em]
                 & RMSE     & 0.680  & 0.332  & \textbf{0.307}  & 0.491  &  0.323     & 0.322  & 0.324$\downarrow$  \\
scm20d           & MAE      & 0.503  & 0.195  & \textbf{0.180}  & 0.352  & 0.188      & 0.187  & 0.187$\uparrow$  \\
                 & NLPD     & 16.568 & 0.543  & 6.882 & 8.682 & 1.341     & 8.097 & \textbf{-0.201}$\uparrow$ \\[0.5em]
                 & RMSE     & 0.391  & \textbf{0.133}  & 0.139  & 0.302  & 0.143      & \textbf{0.133}  & 0.143  \\
WindTurbine      & MAE      & 0.311  & 0.074  & 0.080  & 0.236  & \textbf{0.073}      & 0.074  & \textbf{0.073}  \\
                 & NLPD     & 1.435  & -2.649 & 2.594  & 0.104  & \textbf{-8.749}     & -2.627 & -7.467$\downarrow$ \\[0.5em]
                 & RMSE     & 0.752  &    NA    &   NA     & 0.659       &\textbf{0.547}      &    NA    & 0.556$\downarrow$  \\
Energy           & MAE      & 0.605  &    NA    &     NA   &   0.516     & 0.400      &    NA    & \textbf{0.398}$\uparrow$  \\
                 & NLPD     & 14.775 &    NA    &   NA   &   14.169      & 11.745     &    NA    & \textbf{8.610}$\uparrow$  \\[0.5em]
                 & RMSE     & 0.983  &    NA    &   NA   &  0.955 & \textbf{0.927}      &    NA    &    0.934$\downarrow$\\
usFlight         & MAE      & 0.529  &    NA    &   NA   &  0.494 & \textbf{0.492}      &    NA    &    0.505$\downarrow$\\
                 & NLPD     & 2.331  &    NA    &   NA     &  2.251      & 2.178      &    NA    &    \textbf{2.091}$\uparrow$\\ \bottomrule
\end{tabular}
\caption{Root Mean Square Error (RMSE), Mean Absolute Error (MAE), and Negative Log Predictive Density (NLPD) of state-of-the-art approaches and MOMoGP (our work) on benchmark data sets with 4K to 500K observations. Smaller values are better. The best result is indicated in \textbf{bold} and comparison to the DSMGP is indicated using arrows $\uparrow/\downarrow$. \label{tab:results}}
\end{table*}

In the experiments, we set $K_{S}=2$, $K_{\ProductNode_y}=2$, and $M \in \{500,1000,5000\}$ based on the size of data set. 
Each GP leaf was equipped with a Mat\'ern-$3/2$ covariance function with Automatic Relevance Detection (ARD), and a zero-mean function. 
The initialized lengthscale parameters of the GPs were randomly sampled. 
The learning rate and the number of training epochs were tuned to speed up the training, and at the same time, avoid overfitting. 
That is, when the training loss reached a plateau, the optimisation terminated. We used Adam to maximize the marginal likelihood of the GP leaves. 

The hyperparameters $\theta _{\Leaf}$ of all GPs in our experiments were sampled from a Gamma distribution $\Gamma(2,3)$.
Note that we kept consistent settings for all the GP-related approaches, \eg,~covariance and mean functions. 
For MOSVGP, the number of inducing points was set $Q = 500$, and the number of latent functions corresponded to the number of output dimensions.

Additionally, we tested our hypothesis that a mixture of independent single-output GPs can model correlations between the outputs by employing a shallow mixture of exact single-output GPs denoted as sumGP.
In fact, sumGPs are a sub-class of MOMoGPs which contain only sum nodes, product nodes which split the output space, and GP leaves.

For quantitative evaluation, we compared the Root Mean Squared Error (RMSE):
\begin{equation}
    \text{RMSE}(\mbf{y},\hat {\mbf{y}}) = \frac{1}{P} \sum_{p=1}^{P}\sqrt{\frac{1}{N} \sum_{n=1}^{N}  (y_{n, p} - \hat{y}_{n, p})^2   } \, ,
\end{equation}
the Mean Absolute Error (MAE):
\begin{equation}
    \text{MAE}(\mbf{y},\hat {\mbf{y}}) = \frac{1}{N\cdot P} \sum_{n=1}^{N} \sum_{p=1}^{P} \left | y_{n, p} - \hat{y}_{n, p} \right | \, ,
\end{equation}
and the Negative Log Predictive Density (NLPD):
\begin{equation}
    \text{NLPD}(\mbf{y}_n) = -\log p(\mbf{y}_n \mid \mathfrak{D}, \mbf{x}_n, \theta) \, ,
\end{equation}
where $\hat{\mbf{y}}_n$ is the ground truth of prediction $\mbf{y}_n$ for datum $ \mbf{x}_n$. 

\subsection{Experiment Results}
Herein, we investigate the performance of seven different regression models, including linear regression (LR), exact GP (GP), deep structured mixtures of GPs \mbox{(DSMGP)}~\citep{trapp2020deep},
which all employ independence assumption for the output space, and 
Multitask GP regression \mbox{(MOGP)}~\citep{williams2007multi}, multi-output sparse Variational GP \mbox{(MOSVGP)}~\citep{moreno2018heterogeneous}, sumGP and \mbox{MOMoGP}, which models the output space jointly.
Note that we use consistent settings for all the GP approaches.

\cref{tab:results} reports the RMSE, MAE and NLPD on each data set.
Generally, the multi-output models provide smaller RMSE and MAE values compared with the single-output models.
This means by modelling the joint of the output space, instead of assuming them to be independent, the models achieve a smaller approximation error. 
Moreover, MOMoGP captures predictive uncertainties better than expert-based approaches and DSMGP, resulting in lower NLPDs.
Note that the NLPD gives rise to the output distribution, while the RMSE and the MAE only account for the mean value of the distribution.
Thus, improvements in terms of the NLPD are strictly more important than in terms of the RMSE or the MAE.
Additionally, MOMoGP provided the lowest NLPD values for large-scale data sets such as Energy and usFlight, and the corresponding RMSE and MAE values are also very competitive. 

Overall, we can conclude that MOMoGP can achieve competitive regression results, and provides better predictive uncertainties at the same time.

\begin{figure*}[t]
\setlength{\figurewidth}{0.248\textwidth}
\tiny
\centering
\pgfplotsset{every x tick scale label/.style={at={(rel axis cs:0.9,-0.45)},anchor=south west,inner sep=1pt},scaled ticks=false}
\begin{minipage}[t]{\figurewidth}
    \begin{tikzpicture}[outer sep=0]
    \node[draw=white, fill=black!20, minimum size=\textwidth, inner sep=0pt] (i1) at (0, 0) {\includegraphics[width=\textwidth]{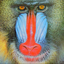}};
    \node[draw=white, thick, fill=black!20, minimum size=0.5\textwidth, inner sep=0pt] (i2) at (-0.25\textwidth, -0.25\textwidth) {\includegraphics[width=0.5\textwidth]{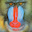}}; 
    \end{tikzpicture}
  \centering{\small{Original and ground truth}}
\end{minipage}
\hfill
\begin{minipage}[t]{\figurewidth}
  \includegraphics[width=\textwidth]{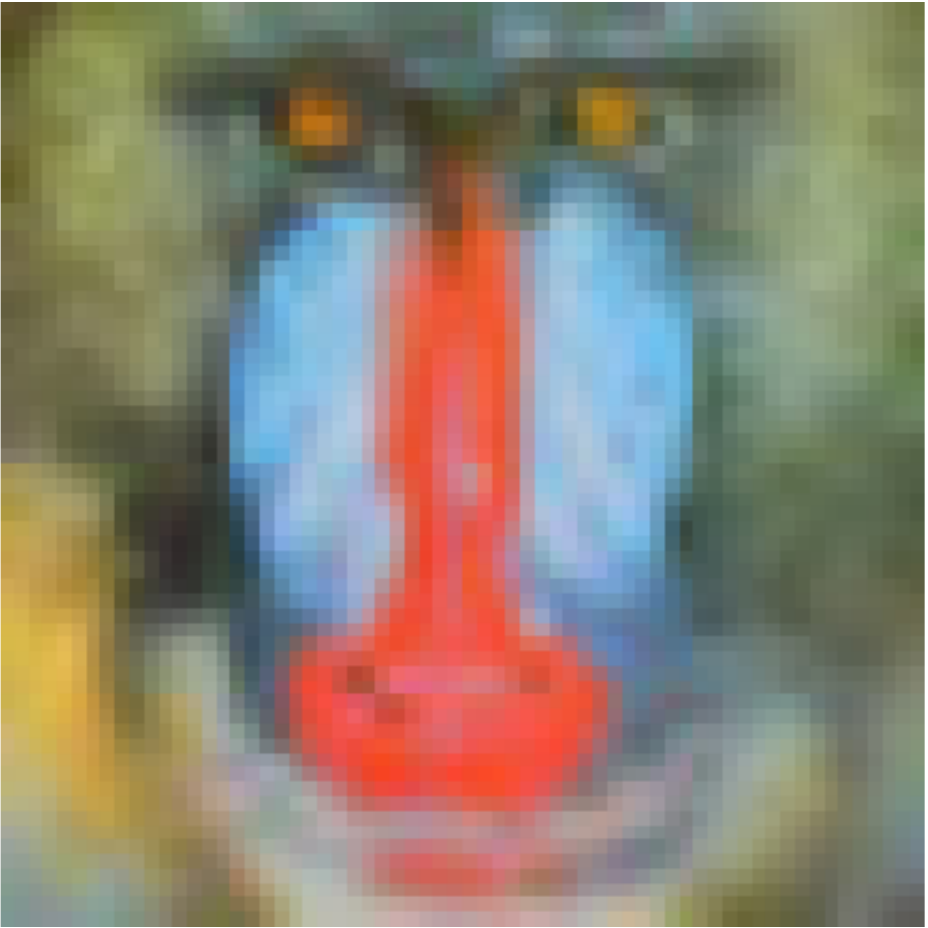}
  \centering{\small{Bilinear reconstruction}}
\end{minipage}
\hfill
\begin{minipage}[t]{\figurewidth}
  \includegraphics[width=\textwidth]{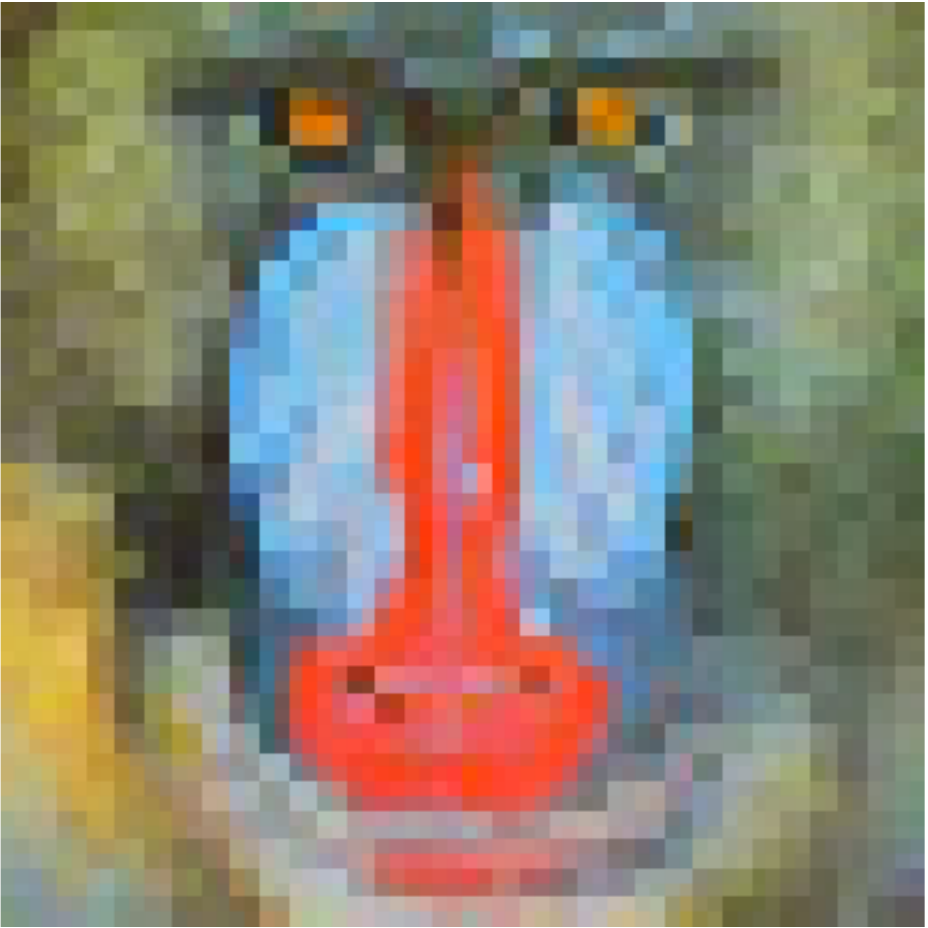}
  \centering{\small{Nearest neighbour reconstruction}}
\end{minipage}
\hfill
\begin{minipage}[t]{\figurewidth}
  \includegraphics[width=\textwidth]{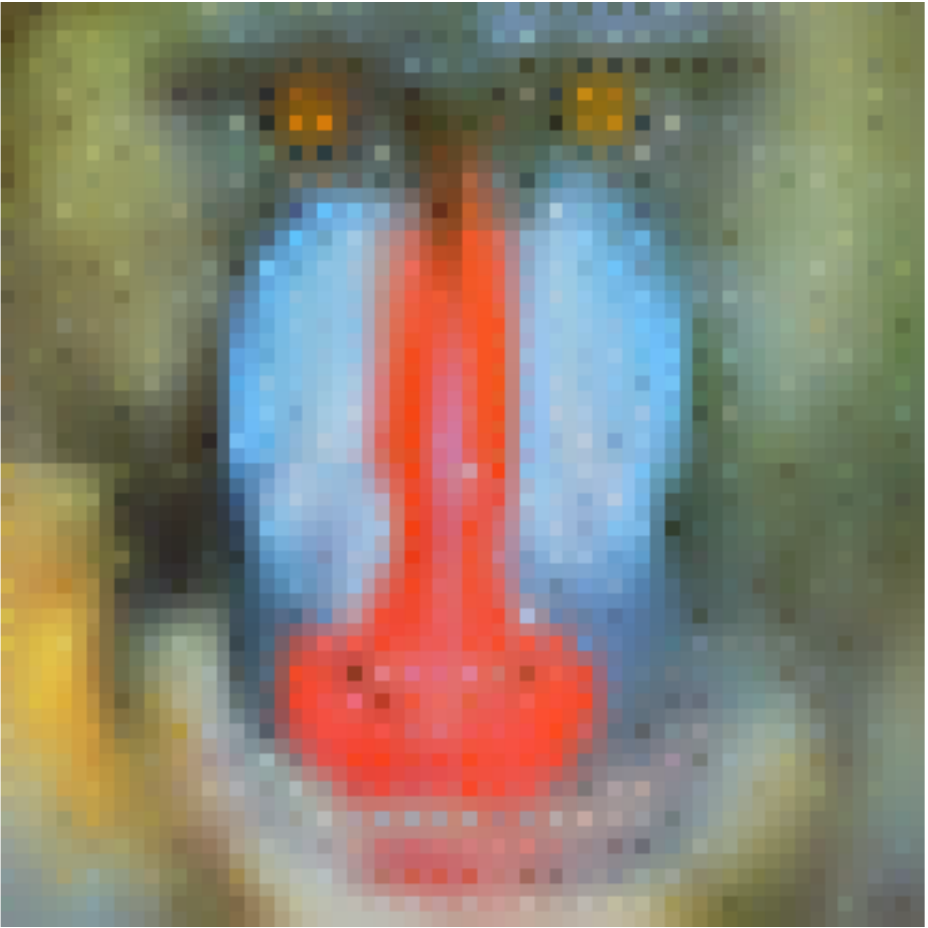}
  \centering{\small{\textbf{MOMoGP reconstruction}}}
\end{minipage}
\caption{Image upsampling using MOMoGP and other methods. MOMoGP obtains the best reconstruction with an RMSE of $1.289$, while Bilinear has RMSE of $1.542$ and Nearest neighbour of $1.855$. (Best viewed in color) \label{Fig.lable}} 
\end{figure*}

\subsection{Extra Results on Image Upsampling}
To deepen the performance evaluation, we envisioned a MOMoGP application to the field of image upsampling. 
The horizontal and vertical locations of a pixel form the input space, while the RGB channels of the pixel are defined as the outputs.
Therefore, we have $D=2$ and $P=3$ for the image upsampling task.
For a given image, this task aims at enlarging the image via interpolation. 
The posterior mean from a MOMoGP is taken as the new pixel value, given the location of the pixel to be interpolated. 
In this experiment, the original image has the size of $64 \times 64$ and was downsampled to $32 \times 32$.
We then aim to reconstruct the original image from the downsampled version.
For MOMoGP, we set $K_{S}=2$, $K_{\ProductNode_x}=2$, $K_{\ProductNode_y}=2$ and $M = 256$.
As visualized in \cref{Fig.lable}, bilinear interpolation produces smooth and blurry artefacts. 
The nearest neighbour approach brings blocks in the image.
MOMoGP as an interpolation approach achieves the best performance, exhibiting a more appropriate balance between colour flattening and salient edge.

\section{Conclusion}
We introduced the multi-output mixture of Gaussian processes (MOMoGP), which leverages a probabilistic circuit with single-output Gaussian process (GP) expert leaves to model correlations between the dependent variables.
%
In comparison to \citep{trapp2020deep} and other expert-based approaches, our approach models the output space jointly, \ie, our model is more general than the approach by \citep{trapp2020deep}, by utilising a recursive decomposition of the output space.
We have shown that this additional decomposition enables MOMoGPs to retain exact posterior inference while also allowing the model to capture dependencies between the outputs without introducing a cubic cost in the number of output dimensions.
In particular, we have shown that we can efficiently approximate the predictive posterior distribution for an unseen datum with its closest multivariate Gaussian distribution through an extension of the approach proposed by \cite{trapp2020deep}.
Finally, we show that MOMoGPs provide competitive results for both RMSE and MAE and often outperform the model by \cite{trapp2020deep} as well as multi-output sparse variational GPs \citep{moreno2018heterogeneous} in terms of the NLPD, indicating that MOMoGPs provide a better estimate of the target distribution.

Our work provides several avenues for future work, \eg, exploiting the spectral representation of stationary GPs \citep{RasmussenW06}, combining MOMoGPs with conditional sum-product networks for mixed multi-output regression \citep{shao2020cspn} and applying MOMoGPs to multi-task Bayesian optimisation \citep{swersky2013bo}.



\begin{acknowledgements} 
The authors thank the anonymous reviewers for their valuable feedback. 
This work was supported by the Federal Ministry of Education and Research (BMBF; project ``MADESI'', FKZ 01IS18043B, and Competence Center for AI and Labour; ``kompAKI'', FKZ 02L19C150), the Hessian Ministry of Higher Education, Research, Science and the Arts (HMWK; projects ``The Third Wave of AI'' and ``The Adaptive Mind''), the Hessian research priority programme LOEWE within the project ``WhiteBox'', and the National Research Center for
Applied Cybersecurity ATHENE, a joint effort of BMBF and HMWK.  
M.T. acknowledges funding from the Academy of Finland (grant number 324345).

\end{acknowledgements}

\bibliography{uai2021-template}

\appendix


\maketitle

\appendix

\section{MOMoGP Structure Construction}
\cref{tab:ci} compares the results obtained from MOMoGPs constructed either using conditional independence tests or using random splitting. 
We construct MOMoGPs using conditional independence tests for the splitting of the output space, we employed the randomized conditional correlation test \cite{strobl2019approximate}. In all experiments, we used a $p$-value of $0.5$. 
We see that the use of a conditional independence test for the structure construction results in an overall improvement of the performance of MOMoGPs with respect to the RMSE and the MAE.

\begin{table}[b]
\centering
\begin{tabular}{ll|cc} \toprule
\multicolumn{2}{c}{Data Set} & Random & CI \\ \midrule
                 & RMSE & \textbf{0.775} & \textbf{0.775}\\
Parkinsons       & MAE & \textbf{0.605} & \textbf{0.605}\\
                 & NLPD & \textbf{2.208} & \textbf{2.208}\\[0.5em]
                 & RMSE & 0.820 & \textbf{0.816} \\
scm20d           & MAE & 0.630 & \textbf{0.627}\\
                 & NLPD & \textbf{11.416} & 11.470\\[0.5em]
                 & RMSE & 0.143 & \textbf{0.137} \\
WindTurbine      & MAE & 0.073 & \textbf{0.072} \\
                 & NLPD & \textbf{-7.467} & -7.478 \\[0.5em]
                 & RMSE & 0.556 & \textbf{0.516}\\
Energy           & MAE & 0.398 & \textbf{0.358} \\
                 & NLPD & \textbf{8.610} & 9.402\\[0.5em]
                 & RMSE & 0.934 & \textbf{0.914} \\
usFlight         & MAE & 0.505 & \textbf{0.485}\\
                 & NLPD & 2.091 & \textbf{2.075} \\ \bottomrule
\end{tabular}
\caption{Comparison of results obtained for MOMoGPs constructed using conditional independence tests (CI) or using random splitting. \label{tab:ci}}
\end{table}

\section{Multi-Output Regression Benchmark Results}
We train and test GP, DSMGP, sumGP, and MOMoGP on all data sets (except for usFlight) five times with different random seeds. 
For usFlight, the above models are trained twice. 
\cref{tab:variance} shows both average and standard deviation of the multiple runs. 
MOGP and MOSVGP have only been trained once for all data sets, thus, their results are not compared in \cref{tab:variance}.

\begin{table*}[t]
\centering
\begin{tabular}{ll|*{4}{rS}} \toprule
\multicolumn{2}{c}{Data Set} & \multicolumn{2}{c}{GP}     & \multicolumn{2}{c}{DSMGP}   & \multicolumn{2}{c}{sumGP}    & \multicolumn{2}{c}{\textbf{MOMoGP}}  \\ \midrule
                 & & mean & \multicolumn{1}{c}{std} & mean   &  \multicolumn{1}{c}{std} & mean &  \multicolumn{1}{c}{std} & mean & \multicolumn{1}{c}{std} \\\midrule
                 & RMSE     & $0.783$  & 0.0004  & \textbf{0.774}  & 0.0001  & $0.784$ & 0.0003  & 0.775$\downarrow$ & 0.0005 \\
Parkinsons       & MAE      & 0.610  & 0.0003  & \textbf{0.604}  & 0.0009  & 0.610 & 0.0004  & 0.605$\downarrow$ & 0.0009 \\
                 & NLPD     & 2.389  & 0.0010  & 2.319  & 0.0030  & 2.388      & 0.0010  & \textbf{2.208}$\uparrow$ & 0.0030 \\[0.5em]
                 & RMSE     & 0.332  & 0.0180  & 0.323  & 0.00027  &  \textbf{0.322}     & 0.00016  & 0.324$\downarrow$ & 0.0011  \\
scm20d           & MAE      & 0.195  & 0.0159  & 0.188  & 0.00041  & \textbf{0.187}      & 0.00032  & \textbf{0.187}$\uparrow$ & 0.0012 \\
                 & NLPD     & 0.543  & 0.0190  & 1.341  & 0.0515  & 8.097     & 0.177  & \textbf{-0.201}$\uparrow$& 0.190  \\[0.5em]
                 & RMSE     & \textbf{0.133}  & 0  & 0.143  & 0.0001  & \textbf{0.133}      & 0.0002  & 0.143 & 0 \\
WindTurbine      & MAE      & 0.074  & 0.0003  & \textbf{0.073}  & 0  & 0.074      & 0.0002  & \textbf{0.073} & 0 \\
                 & NLPD     & -2.649 & 0.0434  & \textbf{-8.749} & 0.0023  & -2.627     & 0.0289  & -7.467$\downarrow$ & 0.0030 \\[0.5em]
                 & RMSE     &   NA   &    NA   &  \textbf{0.547} &   0.0060 &    NA      &    NA    & 0.556$\downarrow$ & 0.0060 \\
Energy           & MAE      &   NA   &    NA   &  0.400 &   0.0020 &    NA      &    NA    & \textbf{0.398}$\uparrow$ & 0.0030 \\
                 & NLPD     &   NA   &    NA   &  11.745&   0.0150 &    NA      &    NA    & \textbf{8.610}$\uparrow$ & 0.0100 \\[0.5em]
                 & RMSE     &   NA   &    NA   &  \textbf{0.927} &  0.0002 &    NA      &    NA    & 0.934$\downarrow$ & 0.0003 \\
usFlight         & MAE      &   NA   &    NA   &  \textbf{0.492} &  0.0003 &    NA      &    NA    & 0.505$\downarrow$ & 0.0003 \\
                 & NLPD     &   NA   &    NA   &  2.178 &  0.0001 &    NA      &    NA    & \textbf{2.091}$\uparrow$ & 0.0357 \\ \bottomrule
\end{tabular}
\caption{Mean and standard deviation of Root Mean Square Error (RMSE), Mean Absolute Error (MAE), and Negative Log Predictive Density (NLPD) of state-of-the-art approaches and MOMoGP (our work) on benchmark data sets. Smaller values are better. Best result is indicated in \textbf{bold} and comparison of MOMoGP to DSMGP is indicated using arrows $\uparrow/\downarrow$. \label{tab:variance}}
\end{table*}


\end{document}